\documentclass[conference]{IEEEtran}
\IEEEoverridecommandlockouts
\usepackage{cite}
\usepackage{svg}
\usepackage{amsmath,amssymb,amsfonts}

\usepackage{algorithmic}
\usepackage{graphicx}
\usepackage{tabularx}
\usepackage{textcomp}
\usepackage{xcolor}
\usepackage{glossaries}
\usepackage{amsmath}
\usepackage{bbm}
\usepackage{url} 
\usepackage{subcaption}
\def\BibTeX{{\rm B\kern-.05em{\sc i\kern-.025em b}\kern-.08em
    T\kern-.1667em\lower.7ex\hbox{E}\kern-.125emX}}
\begin{document}

\newacronym{fss}{FSS}{Few Shot Segmentation}
\newacronym{cv}{CV}{Computer Vision}
\newacronym{iou}{IoU}{Intersection over Union}
\newacronym{miou}{mIoU}{mean Intersection over Union}
\newacronym{dl}{DL}{Deep Learning}
\newacronym{ai}{AI}{Artificial Intelligence}

\title{More than the Sum of Its Parts: Ensembling Backbone Networks for Few-Shot Segmentation\\
{}
\thanks{This study was conducted within the Agritech National Research Center and received funding from the European Union Next-GenerationEU (PIANO NAZIONALE DI RIPRESA E RESILIENZA (PNRR) – MISSIONE 4 COMPONENTE 2, INVESTIMENTO 1.4 – D.D. 1032 17/06/2022, CN00000022). This manuscript reflects only the authors’ views and opinions, neither the European Union nor the European Commission can be considered responsible for them.}
}

\author{
    \IEEEauthorblockN{Nico Catalano\IEEEauthorrefmark{1}, Alessandro Maranelli\IEEEauthorrefmark{2}, Agnese Chiatti\IEEEauthorrefmark{1}, Matteo Matteucci\IEEEauthorrefmark{1}}
    \IEEEauthorblockA{Department of Electronics, Information and Bioengineering\\
    Politecnico di Milano\\
    \IEEEauthorrefmark{1}\{name.surname\}@polimi.it}
    \IEEEauthorblockA{\IEEEauthorrefmark{2}
    \{name.surname\}@mail.polimi.it}
}

\maketitle

\begin{abstract}
Semantic segmentation is a key prerequisite to robust image understanding for applications in \acrlong{ai} and Robotics. \acrlong{fss}, in particular, concerns the extension and optimization of traditional segmentation methods in challenging conditions where limited training examples are available. A predominant approach in \acrlong{fss} is to rely on a single backbone for visual feature extraction. Choosing which backbone to leverage is a deciding factor contributing to the overall performance. In this work, we interrogate on whether fusing features from different backbones can improve the ability of \acrlong{fss} models to capture richer visual features. To tackle this question, we propose and compare two ensembling techniques—Independent Voting and Feature Fusion. Among the available \acrlong{fss} methods, we implement the proposed ensembling techniques on PANet. The module dedicated to predicting segmentation masks from the backbone embeddings in PANet avoids trainable parameters, creating a controlled `in vitro' setting for isolating the impact of different ensembling strategies. Leveraging the complementary strengths of different backbones, our approach outperforms the original single-backbone PANet across standard benchmarks even in challenging one-shot learning scenarios. Specifically, it achieved a performance improvement of +7.37\% on PASCAL-5\textsuperscript{i} and of +10.68\% on COCO-20\textsuperscript{i} in the top-performing scenario where three backbones are combined. These results, together with the qualitative inspection of the predicted subject masks, suggest that relying on multiple backbones in PANet leads to a more comprehensive feature representation, thus expediting the successful application of \acrlong{fss} methods in challenging, data-scarce environments. 
\end{abstract}


\section{Introduction}
Efficient and robust image understanding is a crucial missing capability in Artificial Intelligence (AI) and Robotics that supports key tasks ranging from autonomous driving \cite{bellusci2024semantic,geiger2012we} to precision agriculture \cite{10256348,patricio2018computer,mavridou2019machine} and clinical analysis \cite{esteva2021deep}, to name just a few. One key prerequisite to robust image understanding is the semantic segmentation problem, which involves predicting category labels at the pixel level in a given image \cite{garcia2018survey,guo2018review,8637467}. Following recent AI advancements, various \acrfull{dl} models have been introduced, including U-Net \cite{unet}, Mask R-CNN \cite{maskrcnn},  and PSPNet \cite{pspnet}, that exhibit an impressive performance on popular benchmark datasets in Computer Vision \cite{pascal-voc-2012,MS-COCO,geiger2012we,Cordts2016Cityscapes}. However, these methods share the significant drawback of relying on large-scale training datasets that are expensive to curate. This characteristic of traditional \acrshort{dl} methods for semantic segmentation drastically limits their applicability in scenarios of data scarcity, as well as their ability to generalize beyond the training data distribution. 

To address this limitation and facilitate the widespread adoption of segmentation models in domain-specific applications, the field of  \acrfull{fss} has emerged. In the \acrshort{fss} framework, the objective is to design a model that can learn from limited training examples to accurately segment novel classes as soon as these classes are first observed. Typically, this involves providing the model with as few as one to five labeled examples. 

A proven strategy for implementing \acrshort{fss} involves fine-tuning backbone architectures that have been pre-trained on large-scale, general-purpose datasets, leveraging the benefits of transfer learning. This strategy capitalizes on the diverse and informative features implicitly learned by the backbone during pre-training on a general dataset. A common practice is then to capitalise on the richer features learned on larger-scale sets to adapt the model to a target domain-specific represented by a few training examples. 

In this context, choosing a specific backbone can significantly influence the final performance. Widely adopted backbones, such as VGG \cite{vgg}, ResNet \cite{resnet} and MobileNet \cite{mobilenet} are each characterised by their distinct design and, consequently, provide embeddings that represent different feature sets. To underscore the impact of backbone selection on the overall performance of the model, a common practice in the \acrshort{fss} community is to compare the performance of different backbones. However, the impact of backbone selection on the final performance has not yet been fully studied. In this paper, we build on the intuition that embeddings ensembled from multiple backbones can capture a more comprehensive and descriptive set of image features than those extracted from a single backbone. The underlying expectation is that adopting ensembling strategies will improve the performance of a model on \acrshort{fss}.

To test this hypothesis, we focused on the PANet \cite{panet} \acrshort{fss} model. In PANet, segmentation masks are directly predicted from the backbone embeddings without introducing any trainable parameters. As such, PANet is an ideal candidate for conducting an `in vitro' experiment. Moreover, because PANet is not dependent on mask prediction parameters, it provides a fully modular solution for the integration and evaluation of multiple backbones. That is, results obtained with this setup can be more easily abstracted and extended to different architectures and tasks. Another factor we control for in our experiments is the impact of the dataset chosen for pre-training the different backbones. Namely, we will rely on backbones that have all been pre-trained on ImageNet \cite{deng2009imagenet}. 

These methodological choices allow us to systematically evaluate the effects of leveraging embeddings extracted from different backbones and ensembled with different policies while removing the effects of both the mask prediction stage and the dataset used for pre-training each backbone. Therefore, results obtained in this experimental setup can only be ascribed to the introduction of ensembling strategies.

Additionally, we focus on the more challenging scenario where only one example is provided to the \acrshort{fss} model, putting even more emphasis on the ability of the model to generalize and adapt effectively in a low-data setting. 
Combining multiple backbones in the same pipeline makes particular sense in \acrshort{fss} and one-shot learning scenarios. On the one hand, adding multiple backbones also increases the number of training parameters. However, because \acrshort{fss} methods are trained to generalize to unseen classes, once trained, the model will adapt to an unseen class without requiring additional training, differently from traditional fully-supervised methods.

In sum, in this paper we make the following contributions:
\begin{itemize}
    \item{we present what is, to the best of our knowledge, the first study of ensembling features learned through different backbones for \acrlong{fss}.}
    \item{we devise a series of controlled experiments to disentangle the performance effects exclusively related to ensembling from other contributing effects, namely the impact of the mask prediction module training and the data chosen for pre-training the backbones. This experimental design ultimately facilitates the abstraction of general findings (i.e., applicable to different tasks, models, and domains) from the individual experiments presented in this paper. }
    \item{we demonstrate that ensembling multiple backbones can drastically improve the \acrlong{fss} performance on popular benchmarks datasets, improving up to +7.37\% on PASCAL-5\textsuperscript{i} and up +10.68\% on COCO-20\textsuperscript{i} in terms of average \acrshort{miou}.
    }
\end{itemize}

\section{Background}
Before exploring the application of ensembling techniques for the \acrshort{fss} problem, in this section we thoroughly define the \acrshort{fss} task, as well as the main background concepts related to this task. Concurrently, we illustrate how \acrshort{fss} is approached in the PANet architecture. Subsequently, we delve into the fundamentals of ensembling techniques, laying the ground for a more detailed exploration of their role in the context of \acrshort{fss}.

\subsection{\acrlong{fss}}
Numerous studies in the literature  \cite{BMVC2017_167,rakelly2018conditional,zhang2020sg,zhang2019canet,tian2020prior,liu2020part,wu2021learning,iqbal2022msanet} frame the \acrfull{fss} problem as one of predicting the region mask $\hat{M}_q$ of a subject class $l$ in a query image $I_q$, given a support set $S$ composed of $k$ image-mask pairs. In this context, for a semantic class $l$, the support set 
\[
 S(l) = \{(I^i, M^i_l)\}_{i=1}^k
\]
is the collection of $k$ image-mask pairs that describes the novel class $l$.  
On the other hand, the query image $I_q$ is the image on which the model will predict the segmentation mask $\hat{M_q}$ of the class $l$. 
Then, the learning objective of the \acrshort{fss} model is the function $f_{\theta}$
\[
   \hat{M_q} = f_{\theta}(I_q,S(l)),
\]
which predicts the binary mask $\hat{M}_q$ for the semantic class $l$ in the query image $I_q$ described by the $k$ elements in the support set $S(l)$. In this work, we specifically focus on the case where only one support example is available ($k=1
$), also known as one-shot semantic segmentation. This scenario represents a particular instance of the \acrshort{fss} problem. Thus, the key concepts and definitions introduced also hold in this case. 

A prevalent approach in the \acrshort{fss} field is to adopt a meta-learning framework known as episodic training, which was originally proposed by Vinyals et al.\cite{matching_networks} in the context of one-shot semantic segmentation scenarios. Episodic training, as the name suggests, concerns feeding the learning model with a sequence of ``episodes'' in each of which the model has to learn a new class. Specifically, in each training episode, a label class $l$ is first sampled from the set of training classes $L_{train}$. Then, the model is presented with: i) a support set $S$ of images and mask pairs where only the $l$ class is labelled, and ii) a query image $I_q$ with its corresponding ground truth mask $M_q$. In each episode, the training objective is minimizing the loss between the predicted mask $\hat{M}_q$ and the ground truth mask $M_q$. Similarly, the model performance can be assessed through a series of meta-testing episodes, in which subject classes are selected from $L_{test}$, which contains only examples unseen at training time.

To train and compare \acrshort{fss} models via episodic training, researchers commonly resort to the PASCAL-5\textsuperscript{i} and COCO-20\textsuperscript{i} datasets. These datasets are derived from the well-known PASCAL VOC 2012 \cite{pascal-voc-2012} and MS COCO \cite{MS-COCO} collections of natural images. While the former set includes 20 classes, the latter one covers a wider set of 80 subject classes.

PASCAL-5\textsuperscript{i} and COCO-20\textsuperscript{i} are organized into four folds so that the label set $L$ of all subject classes in each original dataset is partitioned into four subsets. For each split, the union of three label subsets will form the $L_{train}$ set used to sample training episodes, while the remaining subset will serve as $L_{test}$ and is thus devoted to sampling testing episodes. In this way are constructed four folds, each one with different $L_{test}$. The two datasets PASCAL-5\textsuperscript{i} and COCO-20\textsuperscript{i} are widely adopted in the literature, and their division into four folds has become a standard practice. Given their prevalence, it is customary to report performance metrics for each of the four folds individually and then provide an average across the folds to summarize the results. In line with this common practice, we follow the same approach, as detailed in Section \ref{section_experimental_results} and exemplified in Tables \ref{tab_pascal} and \ref{tab_coco}.

\subsection{PANet}

\begin{figure*}[t!]
  \centering
\includegraphics[width = 500pt]
{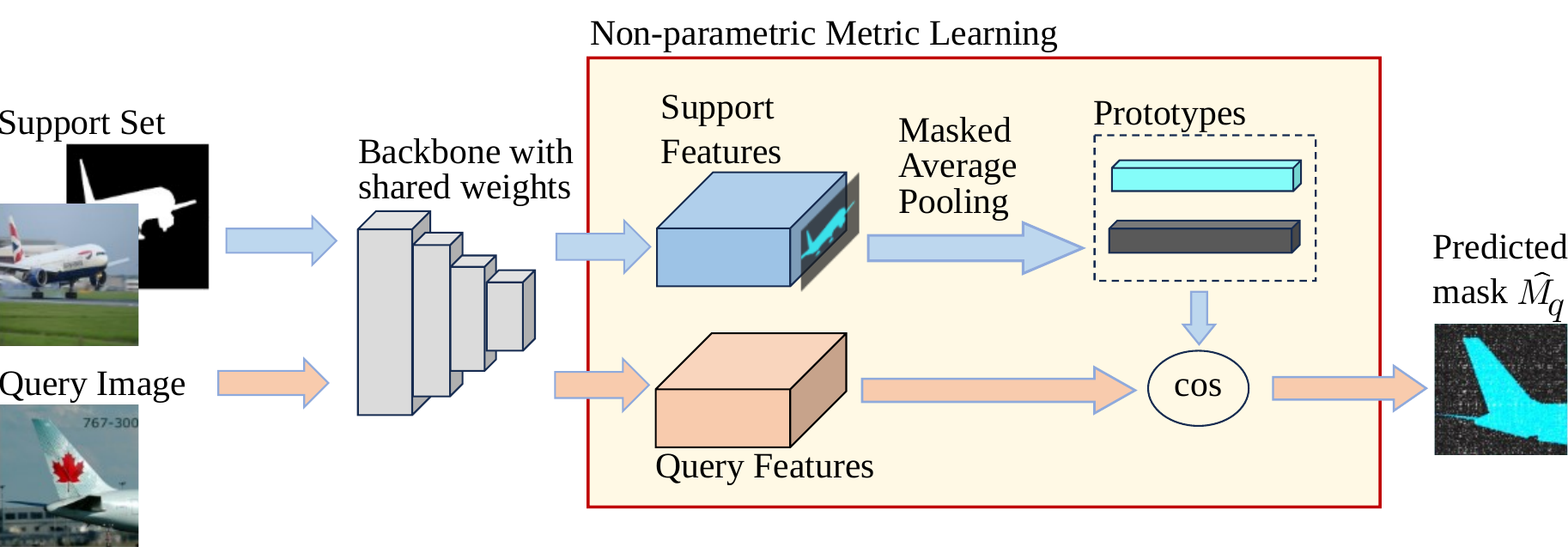}
  \caption{Adapted from \cite{panet}, this diagram illustrates the inference process of PANet. First, features are extracted from both the Query Image and the examples in the Support Set through a shared backbone. Then, Masked Average Pooling is applied to features extracted from the support images, generating prototypes for each labeled subject class. Ultimately, the cosine distance is computed between the embeddings at each spatial location within the query feature volume and each prototype, yielding the predicted mask $\hat{M}_q$.}
\label{fig:panet}
\end{figure*}

PANet \cite{panet} addresses the \acrshort{fss} task through a metric learning approach, where class-specific prototypes are derived from the embeddings of images in the support set. Namely, in each episode, reference embeddings for the support and query images are extracted from a shared backbone network. Subsequently, masked average pooling \cite{zhang2020sg} is applied to the support set embeddings and corresponding masks, yielding a compact prototypical representation of the novel class. The query image is projected into the same feature space as the generated embeddings. Lastly, image segmentation is performed by matching embeddings with the learned prototypes at each pixel location. 
An overview of the inference process of PANet is depicted in Fig.\ref{fig:panet}.

In mathematical terms, given a support set $S_{l} = \{(I_{l,k} , M_{l,k} )\}$ as input, PANet first computes a feature map $F_{l,k}$ for the image $I_{l,k}$. Here, $l = 1, . . . , L$ indexes the label class from the set $\mathcal{C}$ of $L$ considered label classes and $k = 1, . . . , K$ indexes the support image. The set $\mathcal{C}$ can correspond to either $L_{train}$ or $L_{test}$ depending on whether the model is in training or testing mode.  From $F_{l,k}$, the prototype of label class $l$ is computed via masked average pooling \cite{zhang2020sg} as follows:

\begin{equation}
p_l = \frac{1}{K} \sum_{k = 1}^{K} \frac{\sum_{x,y} F_{l,k}^{(x,y)} \mathbbm{1} [ M_{l,k}^{(x,y)}= l]}{\sum_{x,y} \mathbbm{1} [ M_{l,k}^{(x,y)}= l]},
\end{equation}

where $(x, y)$ are the spatial locations and $\mathbbm{1}(·)$ is an indicator function that always outputs $1$ if the argument is true and $0$ otherwise. Similarly, the prototype representation of the background is computed by:

\begin{equation}
p_{bg} = \frac{1}{L*K} \sum_{l \in \mathcal{C}}\sum_{k = 1}^{K}\frac{\sum_{x,y} F_{l,k}^{(x,y)} \mathbbm{1} [ M_{l,k}^{(x,y)} \notin \mathcal{C}]}{\sum_{x,y} \mathbbm{1} [ M_{l,k}^{(x,y)} \notin \mathcal{C}]}.
\end{equation}

The semantic segmentation task can be seen as classification at each spatial location. Thus, PANet computes the cosine distance between the query feature vector and each computed prototype at each spatial location. Then, it applies a softmax operation to the distances to produce a probability distribution $\hat{M}_q$ over the target classes, including the background class.
Let $cos$ be the cosine distance function, $\mathcal{P}=\{p_l | l \in C\} \cup \{p_{bg}\}$ the set of all prototypes, and $F_{q}$ the query feature volume. Then, for each subject class $p_{j} \in \mathcal{P}$ the probability map at the spatial location $(x,y)$ is defined as:
\begin{equation}
    \tilde{M}^{(x,y)}_{q;j} = \frac{{\exp(-cos(F^{(x,y)}_{q}, p_{j}))}}{{\sum_{p_{j} \in \mathcal{P}} \exp(-cos(F^{(x,y)}_{q}, p_{j}))}}.
\end{equation}
Finally, the predicted segmentation mask is obtained by selecting the class index of highest probability for each spatial location:
\begin{equation}
   \hat{M}^{(x,y)}_{q} = \arg\max_{j} \tilde{M}^{(x,y)}_{q;j}. 
\end{equation}

\subsection{Bayesian Voting}
Among ensembling methods, Bayesian Voting \cite{dietterich2000ensemble} is a technique that leverages a probabilistic framework to combine the predictions of multiple base classifiers. It builds on the premise of modelling the classification problem through a Bayesian perspective.

Given a dataset \(X\) with \(N\) samples and corresponding labels \(y\), let \(h_1(x), h_2(x), ..., h_B(x)\) represent the predictions of \(B\) base classifiers on an input instance \(x\). Each base classifier provides a probability distribution over the possible classes for \(x\). The Bayesian Voting process combines these probability distributions to derive the final class probabilities for \(x\). The final predicted class label is often determined by selecting the class with the highest probability:

\begin{equation}
P(y | x) = \sum_{b=1}^{B} c_b P_b(y | x),
\end{equation}

where \(P_b(y | x)\) is the probability distribution given by the \(b\)-th base classifier for class \(y\) on input \(x\), $c_b$ is the multiplicative coefficient for the \(b\)-th base classifier, and \(P(y | x)\) is the final combined probability distribution.

\section{Related Work}
Ensembling methods, involve strategically combining multiple individual models to enhance predictive performance. While the importance of selecting methods with complementary strengths has been emphasized by Dietterich et al. \cite{dietterich2000ensemble}, who highlighted that an effective ensemble relies on accurate individual predictors making errors in different regions of the input space, the application of ensembling to \acrshort{fss} remains relatively unexplored.

In the realm of semantic segmentation, ensembling can be implemented as the combination of multi-scale feature sets generated by feature pyramid network methods \cite{lin2017feature} and fed into independent decoders, creating an ensemble, as explored by Bousselham et al. \cite{Bousselham_2022_BMVC}. Additionally, Khirodkar et al. \cite{khirodkar2022sequential} proposed an ensembling chain where each model is conditioned on both the input image and the prediction of the previous model, allowing each model in the chain to correct the error of the previous.

In the context of Few-Shot Classification, Dvornik et al. \cite{Dvornik_2019_ICCV} applied ensembling by combining different Convolutional Networks trained to produce a single output prediction. 

While ensembling methods have demonstrated promise in related fields such as few-shot classification and semantic segmentation, their application to \acrshort{fss} has been rather limited. To the best of our knowledge, only Yang et al. \cite{yang2020prototype} have explored ensembling methods in the context of \acrshort{fss}. In their work, they addressed the inadequacy of single prototypes per semantic class in \acrshort{fss} by learning multiple prototypes per class, presenting a form of ensembling as the final predictions require integrating multiple probability maps for the same subject class. However, while Yang et al. ensembled the multiple prototypes of a class, our exploration focuses on ensembling different embeddings from the same image.

\section{Methods}
This section describes the experimental methodology we followed to investigate the utility of introducing ensembling techniques in the context of \acrshort{fss} pipelines. Motivated by the limited exploration of ensembling methods in \acrshort{fss}, we specifically focus on the PANet architecture with embeddings produced by different backbones: VGG16, ResNet50, and MobileNet-V3-Large. We chose PANet among the many \acrshort{fss} methods precisely because of the lack of trainable parameters in the model component responsible for processing embeddings and predicting masks. This configuration allows us to isolate the impact of different ensembling strategies by directly examining the evaluation metrics while preventing the model from learning any implicit properties of the new latent space. In particular, we focus our experimentation on the challenging one-shot scenario, which requires to generalize from a single example per class. We consider two distinct ensembling methods: Independent Voting, and Feature Volume Fusion, which we illustrate in more detail in the remainder of this section.

\subsection{Independent Voting}
In this approach, diverse backbones generate independent probability maps, and these maps are then aggregated to form a unified prediction, akin to the principles of Bayesian Voting \cite{dietterich2000ensemble}. This strategy aims to maintain the autonomy of each model during both training and inference, while enabling the combination of features learned through different methods to produce the final predictions.

\begin{figure}[b]
  \centering
\includegraphics[ width = 250pt]{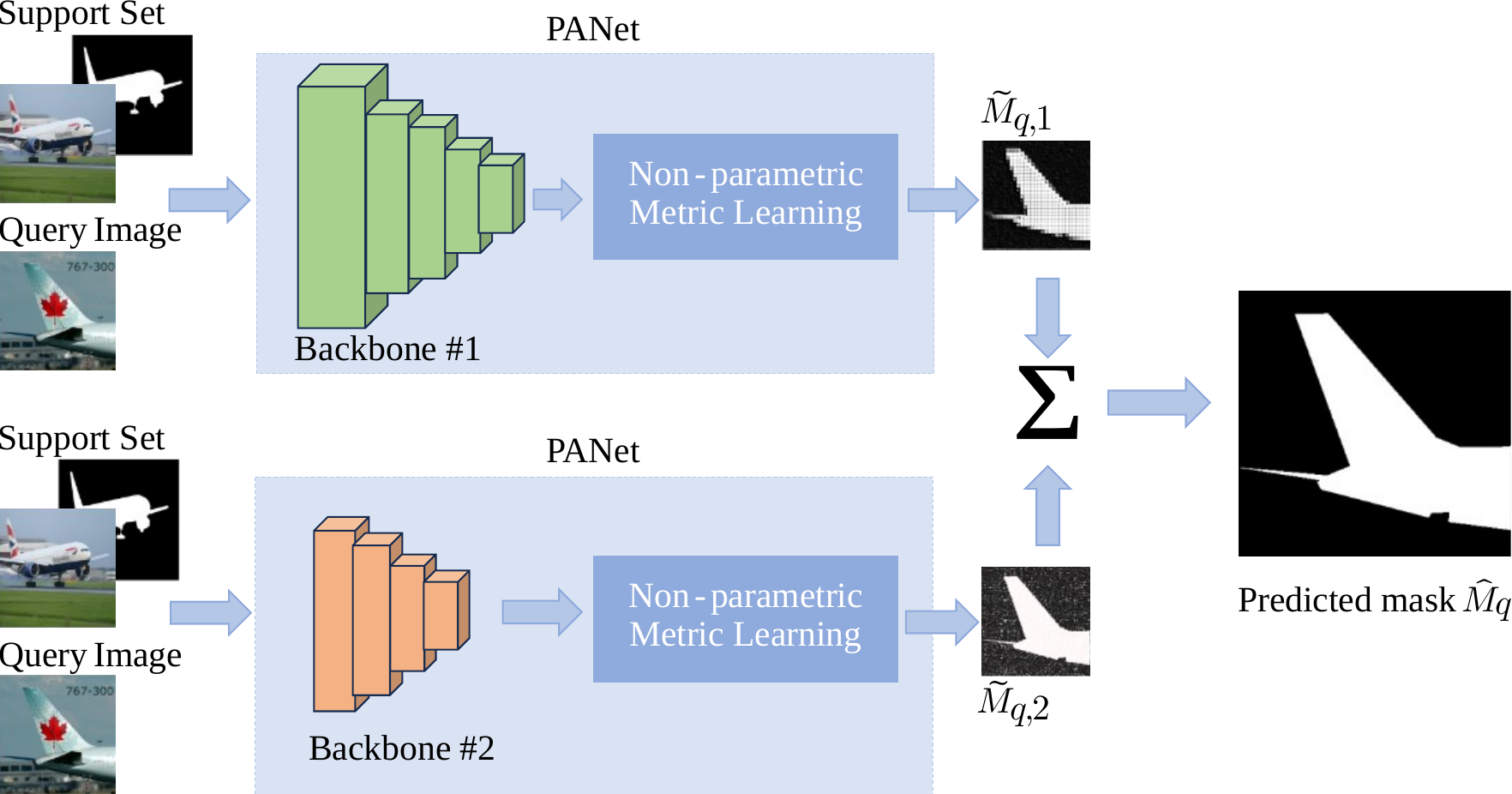}
  \caption{Independent Voting: the Query Image and Support Set examples are passed in parallel through multiple PANet branches, each employing a distinct backbone. The individual probability maps generated by each branch are then combined using Bayesian voting to produce the final prediction, $\hat{M}_q$.}
\label{fig:voting}
\end{figure}

In our implementation, we extract the probability map from each backbone before the application of the softmax function. These individual probability maps are then combined to generate a comprehensive probability map, and the prediction of the ensemble is subsequently derived by applying the softmax function to this combined probability map. A schematic representation of this process is presented in Fig. \ref{fig:voting}.
We opted to ensure an equal contribution from each backbone in the ensemble assigning fixed and equal weights, setting them as the inverses of the number of available backbones. 

The ensemble probability map $\tilde{M}_{q}$ for each subject class is computed accordingly:

\begin{equation}
    \tilde{M}^{(x,y)}_{q;j} = \sum_{b \in \mathcal{B}} \frac {1} {|\mathcal{B}|}  \frac{{\exp(- d(F^{(x,y)}_{q,\text{b}}, p_{j,\text{b}}))}}{{\sum_{p_{j} \in \mathcal{P_{\text{b}}}}\exp(-d(F^{(x,y)}_{q,\text{b}}, p_{j}))}}, 
\end{equation}

where $\mathcal{B}$ represents the set of all backbones, $\mathcal{P}_b=\{p_{c,b} | c \in C_{i,b}\} \cup \{p_{bg,b}\}$ is the set of all prototypes for the backbone $b$, $F_{q,b}$ denotes the query feature map for backbone $b$, and $|\mathcal{B}|$ is the number of involved backbones.

\subsection{Feature Volume Fusion}
\begin{figure}[b]
  \centering
  \includegraphics[width = 250pt]{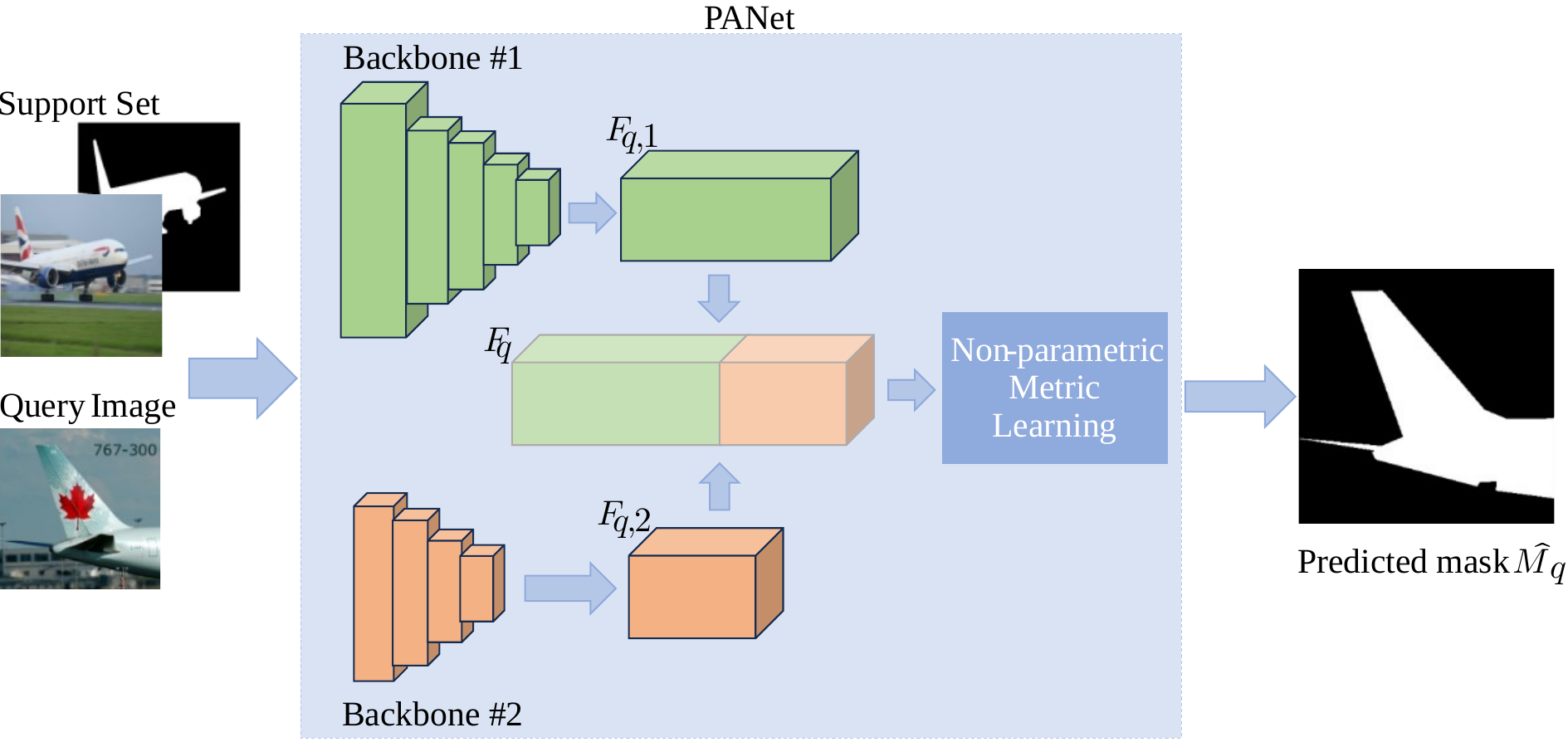}
  \caption{Feature Volume Fusion: This diagram illustrates the Feature Volume Fusion process, where two or more backbones are applied for extracting features from the Query Image and examples in the Support Set. These features are then concatenated along the channel axis, forming a consolidated ensembled feature map. The ensembled feature map is subsequently given as an input to the non-parametric Metric Learning stage of PANet. }
\label{fig:fusion}
\end{figure}

Feature Volume fusion is applied for concatenating multiple backbone embeddings. After passing input images through the backbones, the feature volumes produced by each backbone are concatenated to be later parsed by the Non-parametric metric learning module of PANet. This method is depicted in Fig. \ref{fig:fusion}. The concatenation operation is functional to producing richer feature volumes, comprehensively capturing all the available features extracted on the input data by the backbones and happens as follows:

\begin{equation}
F^{(x,y)}_{q} =  F^{(x,y)}_{q,\text{b1}} || F^{(x,y)}_{q,\text{b2}} || \text{...} || F^{(x,y)}_{q,\text{bn}},
\end{equation}

where $F^{(x,y)}_{q,\text{b}_i}$ is the feature volume produced by the backbone $b_i$ at the spatial location $(x,y)$, and the symbol $||$ denotes the concatenation of feature maps over the channel axis.

\subsection{Evaluation Metrics}
To comprehensively assess the segmentation quality of PANet under different ensembling configurations, we employed two key metrics: the \acrfull{iou} and the \acrfull{miou}. 
The \acrshort{iou} is the ratio of the intersection area between predicted and ground truth masks to their union area. It provides a quantifiable measure of the overlap between the predicted and ground truth masks, with values ranging from 0 to 1, 1 indicating a perfect match.

Viewing segmentation as the problem of classifying individual image pixels, we define $TP_c$, $TN_c$, $FP_c$, and $FN_c$ as the counts of true positives, true negatives, false positives, and false negatives,  predictions for the class $c$ at the pixel level. 
The \acrshort{iou}, specific to class $c$, is then computed as:

\begin{equation} \label{eq:IoU_l}
    IoU_c= \frac{TP_c}{TP_c+FP_c+FN_c}.
\end{equation}

To gauge the overall segmentation performance across the multiple classes in a data fold, we also keep track of the \acrshort{miou}, representing the average \acrshort{iou} value across all object classes in a fixed fold. Mathematically, \acrshort{miou} is expressed as:

\begin{equation} \label{eq:mIou}
    mIoU = \frac{1}{C}\sum_{c=1}^{C}{IoU_c},
\end{equation}  

where $c$ assumes all the index values of a subject class.

Considering that testing a model on a single fold entails testing it on a subset of the labeled classes in the dataset, the \acrshort{miou}, if considered in isolation, may not adequately reflect the overall model performance. Indeed, different backbones and ensembling methods may perform differently on different classes, i.e., on different data folds. Thus, to accompany this metric with global figures that summarise results across different classes and folds, we also consider the mean of the \acrshort{miou} across the four folds, reported in the last row of Tables \ref{tab_pascal} and \ref{tab_coco}. This metric offers a comprehensive overview by averaging the model performance across all folds of the dataset. Consequently, the mean of \acrshort{miou} across the four folds enables the fair comparison of the different models and configurations tested in this study.

\begin{table*}
    \centering
    \caption{Results on PASCAL-5\textsuperscript{i}.}
    \begin{tabularx}{\textwidth}
{     >{\centering\arraybackslash}X |
    | >{\centering\arraybackslash}X 
    | >{\centering\arraybackslash}X 
    | >{\centering\arraybackslash}X |
    | >{\centering\arraybackslash}X 
    | >{\centering\arraybackslash}X 
    | >{\centering\arraybackslash}X 
    | >{\centering\arraybackslash}X |
    | >{\centering\arraybackslash}X 
    | >{\centering\arraybackslash}X 
    | >{\centering\arraybackslash}X 
    | >{\centering\arraybackslash}X 
    }
    \hline
        ~ & \multicolumn{3}{|c||}{Baseline} & \multicolumn{4}{|c||}{Indipendent Voting} & \multicolumn{4}{|c}{Feature Volume Fusion} \\ \hline
        \textbf{} & VGG16 & ResNet50 & MobileNet & VGG16 + ResNet50 & ResNet50 + MobileNet & VGG16 + MobileNet & VGG16 + ResNet50 + MobileNet & VGG16 + ResNet50 & ResNet50 + MobileNet & VGG16 + MobileNet & VGG16 + ResNet50 + MobileNet \\ \hline
        
\textbf{Fold 0}     & 0.4075        & 0.4069 & 0.4557 & 0.4447 & 0.4625 & 0.4587 & 0.4605 & 0.4381 & 0.4293 & 0.4320 & 0.4405 \\ \hline
\textbf{Fold 1}     & 0.5751        & 0.5667 & 0.5578 & 0.6012 & 0.5875 & 0.5928 & 0.6043 & 0.5857 & 0.5835 & 0.5824 & 0.5870 \\ \hline
\textbf{Fold 2}     & 0.5053        & 0.5005 & 0.4752 & 0.5332 & 0.5398 & 0.5266 & 0.5466 & 0.5175 & 0.5202 & 0.5092 & 0.5219 \\ \hline
\textbf{Fold 3}     & 0.4108        & 0.3984 & 0.4056 & 0.4215 & 0.4159 & 0.4236 & 0.4274 & 0.4136 & 0.4063 & 0.4167 & 0.4169 \\ \hline \hline
\textbf{Mean}       & \bf{0.4747}   & 0.4681 & 0.4736 & 0.5002 (+5.36\%) & 0.5014 (+5.63\%) & 0.5004 (+5.41\%) & \bf{0.5097} (+7.37\%) & 0.4887 (+2.95\%) & 0.4848 (+2.13\%) & 0.4851 (+2.18\%) & \bf{0.4916} (+3.56\%) \\ \hline
    \end{tabularx}
    \label{tab_pascal}
\end{table*}

\begin{table*}
    \centering
    \caption{Results on COCO-20\textsuperscript{i}.}
    \begin{tabularx}{\textwidth}
{     >{\centering\arraybackslash}X |
    | >{\centering\arraybackslash}X 
    | >{\centering\arraybackslash}X 
    | >{\centering\arraybackslash}X |
    | >{\centering\arraybackslash}X 
    | >{\centering\arraybackslash}X 
    | >{\centering\arraybackslash}X 
    | >{\centering\arraybackslash}X |
    | >{\centering\arraybackslash}X 
    | >{\centering\arraybackslash}X 
    | >{\centering\arraybackslash}X 
    | >{\centering\arraybackslash}X 
    }
    \hline
        ~ & \multicolumn{3}{|c||}{Baseline} & \multicolumn{4}{|c||}{Indipendent Voting} & \multicolumn{4}{|c}{Feature Volume Fusion} \\ \hline
        \textbf{} & VGG16 & ResNet50 & MobileNet & VGG16 + ResNet50 & ResNet50 + MobileNet & VGG16 + MobileNet & VGG16 + ResNet50 + MobileNet & VGG16 + ResNet50 & ResNet50 + MobileNet & VGG16 + MobileNet & VGG16 + ResNet50 + MobileNet \\ \hline
\textbf{Fold 0} & 0.2849 & 0.2900 & 0.2684  & 0.3149 & 0.3065 & 0.3144 & 0.3254 & 0.3214 & 0.2908 & 0.3140 & 0.3315 \\ \hline
\textbf{Fold 1} & 0.2072 & 0.2212 & 0.1871  & 0.2342 & 0.2291 & 0.2261 & 0.2390 & 0.2343 & 0.2214 & 0.2213 & 0.2394 \\ \hline
\textbf{Fold 2} & 0.1889 & 0.2267 & 0.2077  & 0.2229 & 0.2306 & 0.2260 & 0.2405 & 0.2226 & 0.2345 & 0.2322 & 0.2383 \\ \hline
\textbf{Fold 3} & 0.1521 & 0.1535 & 0.1459  & 0.1662 & 0.1650 & 0.1712 & 0.1750 & 0.1688 & 0.1670 & 0.1667 & 0.1775 \\ \hline \hline
\textbf{Mean}   & 0.2083 & \bf{0.2229} & 0.2023  & 0.2346 (+5.25\%) & 0.2328 (+4.46\%) & 0.2344 (+5.19\%) & \bf{0.2450} (+9.91\%) & 0.2368 (+6.24\%) & 0.2284 (+2.50\%) & 0.2336 (+4.80\%) & \bf{0.2467} (+10.68\%) \\ \hline

    \end{tabularx}
    \label{tab_coco}
\end{table*}

\section{Experiment Results}
\label{section_experimental_results}
Our experimental evaluation, detailed in Tables \ref{tab_pascal} and \ref{tab_coco}, provides valuable insights into the segmentation performance of PANet on the PASCAL-5\textsuperscript{i} and COCO-20\textsuperscript{i} datasets. Initially, we contrasted baseline methods utilizing individual backbones—VGG16, ResNet50, and MobileNet-V3-Large. The \acrshort{miou} scores averaged across folds were consistent across all backbones, with VGG16 and ResNet50 emerging as the best-performing baselines for PASCAL-5\textsuperscript{i} and COCO-20\textsuperscript{i}.

Subsequently, Independent Voting and Feature Volume Fusion were applied for combining two and three backbones and evaluated against the best-performing baselines for each dataset. 

Independent Voting demonstrated significant improvements, achieving a 5\% mean \acrshort{miou} increase for PASCAL-5\textsuperscript{i} with respect to the top-performing baseline for this dataset VGG16, and over 4\% improvement for COCO-20\textsuperscript{i} compared to the best-performing baseline for this dataset ResNet50. The most notable results were obtained when combining all three backbones, with Independent Voting achieving a 7.37\% improvement for PASCAL-5\textsuperscript{i} and 9.91\% for COCO-20\textsuperscript{i} compared to the respective best-performing baselines.

Feature Volume Fusion, applied to pairs of backbones, exhibited improvements of up to 2.95\% for PASCAL-5\textsuperscript{i} over the best-performing baseline for this dataset (VGG16) and 6.24\% for COCO-20\textsuperscript{i} over the best-performing baseline for this dataset (ResNet50). Integration across all three backbones further increased the performance, with a 3.56\% mean \acrshort{miou} increase for PASCAL-5\textsuperscript{i}, and 10.68\% for COCO-20\textsuperscript{i} over their respective best-performing baselines.

Overall, ensembling different backbones consistently resulted in improved metrics. Pipelines relying on the combination of three backbones invariably outperformed methods that ensemble only two feature vectors. This performance trend could be linked to each backbone capturing different feature sets. Once combined, these complementary feature sets can lead to a more informative description of a given image. 

Moreover, it is worth noting how performance differences between different ensembling strategies are influenced by the specific dataset considered for the evaluation. Indeed, Independent Voting demonstrate a clear superiority on PASCAL-5\textsuperscript{i}, while results on COCO-20\textsuperscript{i} indicated a less notable difference between ensembling strategies. Specifically, when using two backbones on COCO-20\textsuperscript{i}, either Independent Voting or Feature Volume Fusion provided a higher performance depending on the specific backbone combination being considered. However, these performance differences were not remarkable, especially when considering the mean IoU across different data folds. 

A dataset-dependent trend in the effectiveness of each strategy can be similarly observed when combining all three backbones. For PASCAL-5\textsuperscript{i} experiments, Independent Voting was the most effective ensembling strategy, surpassing Feature Volume Fusion by 0.0181 points on the mean of \acrshort{miou} across folds. This difference in scores corresponds to a 3.81\% delta in the improvement rate relative to the best-performing backbone. Conversely, for COCO-20\textsuperscript{i}, Feature Volume Fusion outperformed Independent Voting, albeit with a slight margin of only 0.0017 points on the mean of \acrshort{miou} across folds. This difference corresponds to a 0.76\% variation over the improvement rate compared to the best-performing backbone.  

\begin{figure*}[]
    \captionsetup{justification=centering, margin=1cm}
    \centering
    \begin{tabularx}{\textwidth}{@{}*{6}{>{\centering\arraybackslash}X}@{}}
        (a) Ground Truth  &
        (b) MobileNet &
        (c) VGG16  &
        (d) ResNet50  &
        (e) Independent Voting  &
        (f) Feature Volume Fusion  \\
      \includegraphics[width=\linewidth]{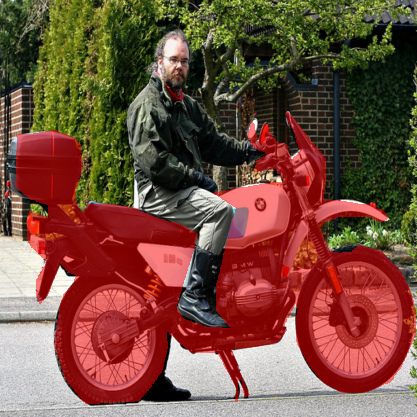} &
      \includegraphics[width=\linewidth]{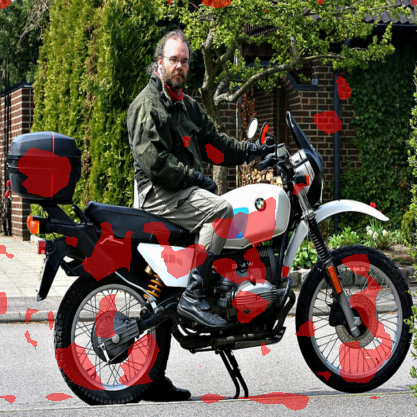} &
      \includegraphics[width=\linewidth]{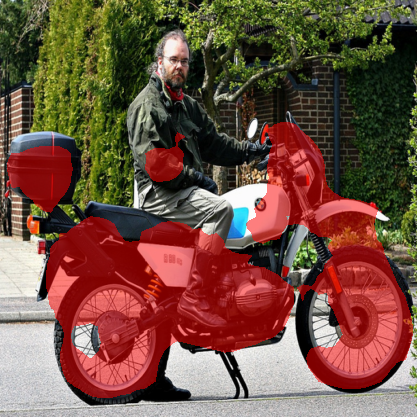} &
      \includegraphics[width=\linewidth]{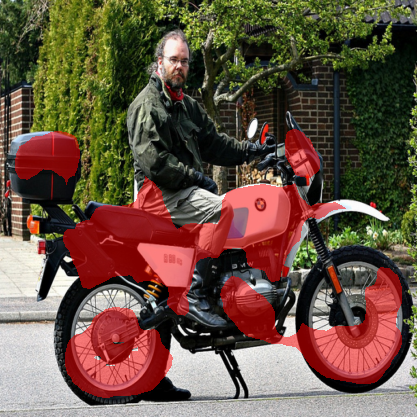} &
      \includegraphics[width=\linewidth]{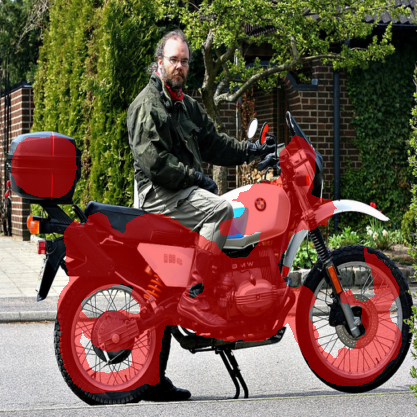} &
      \includegraphics[width=\linewidth]{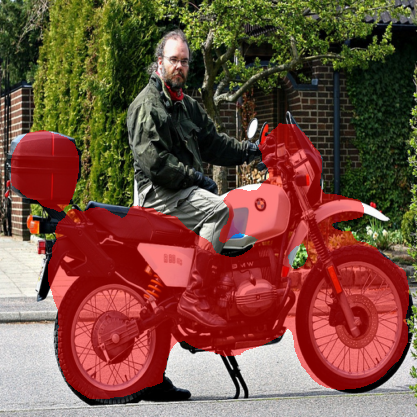} \\
      &
      0.22 \acrshort{iou} &
      0.64 \acrshort{iou} &
      0.51 \acrshort{iou} &
      0.58 \acrshort{iou} &
      0.80 \acrshort{iou} \\
      \includegraphics[width=\linewidth]{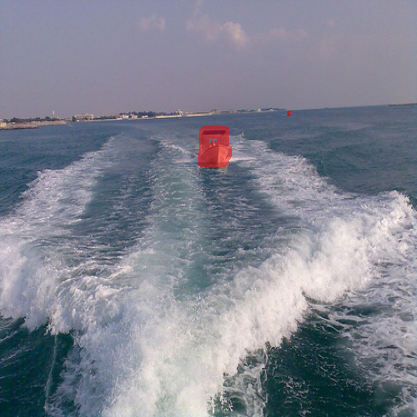} &
      \includegraphics[width=\linewidth]{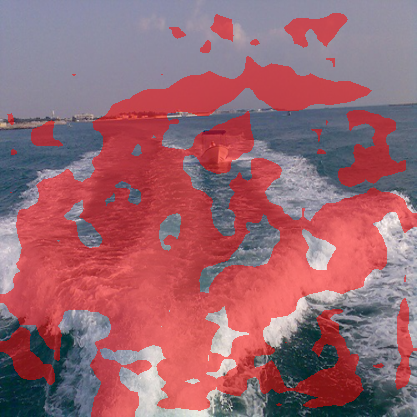} &
      \includegraphics[width=\linewidth]{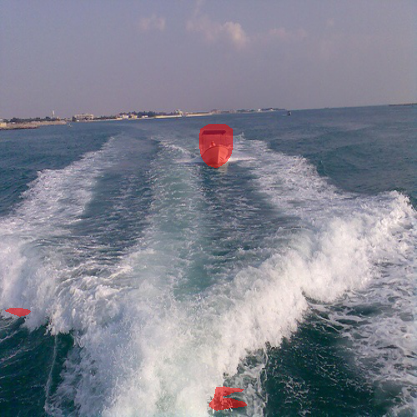} &
      \includegraphics[width=\linewidth]{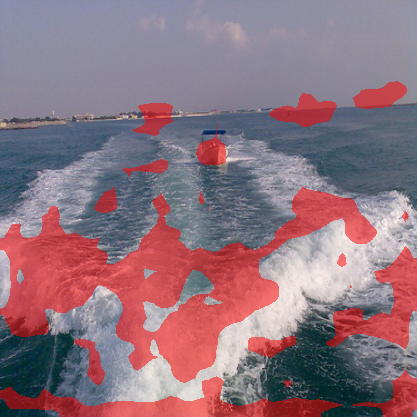} &
      \includegraphics[width=\linewidth]{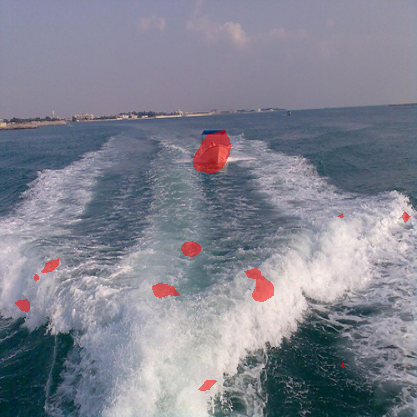} &
      \includegraphics[width=\linewidth]{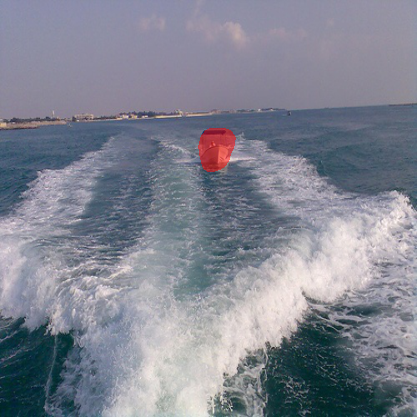} \\
      &
      0.01 \acrshort{iou} &
      0.37 \acrshort{iou} &
      0.01 \acrshort{iou} &
      0.19 \acrshort{iou} &
      0.71 \acrshort{iou} \\
      \includegraphics[width=\linewidth]{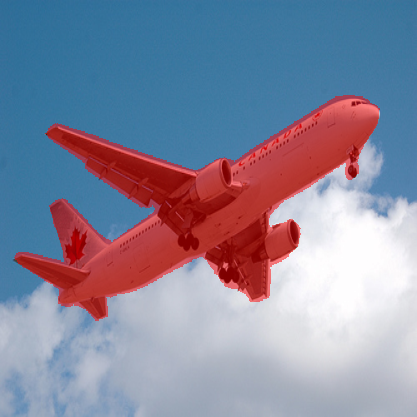} &
      \includegraphics[width=\linewidth]{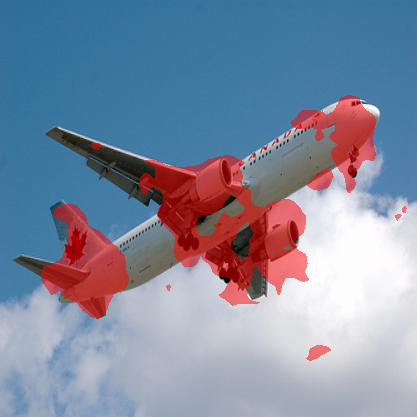} &
      \includegraphics[width=\linewidth]{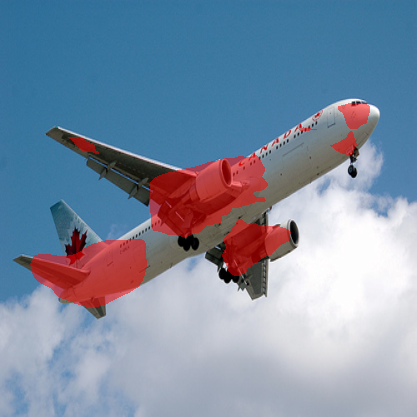} &
      \includegraphics[width=\linewidth]{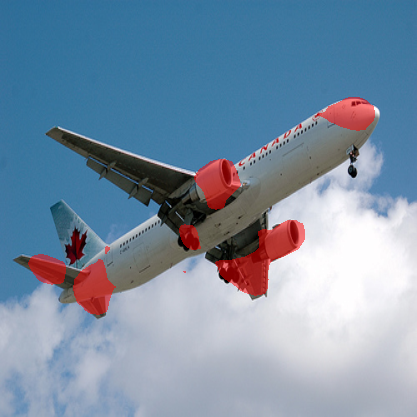} &
      \includegraphics[width=\linewidth]{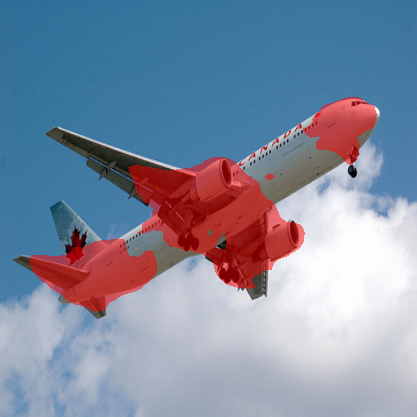} &
      \includegraphics[width=\linewidth]{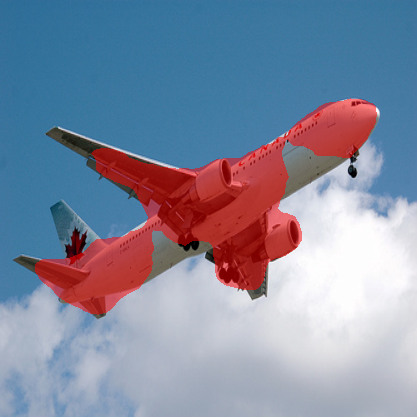} \\
      &
      0.49 \acrshort{iou} &
      0.42 \acrshort{iou} &
      0.23 \acrshort{iou} &
      0.59 \acrshort{iou} &
      0.70 \acrshort{iou} \\
      \includegraphics[width=\linewidth]{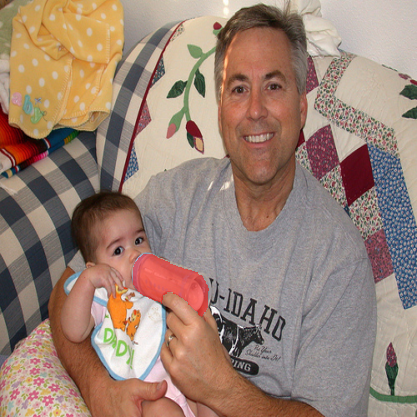} &
      \includegraphics[width=\linewidth]{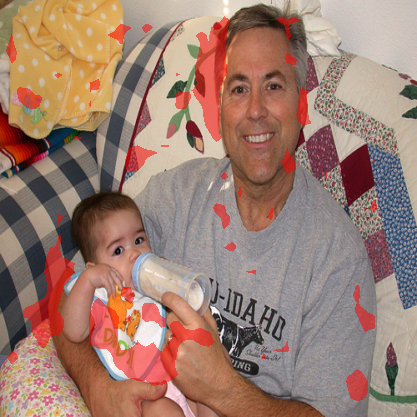} &
      \includegraphics[width=\linewidth]{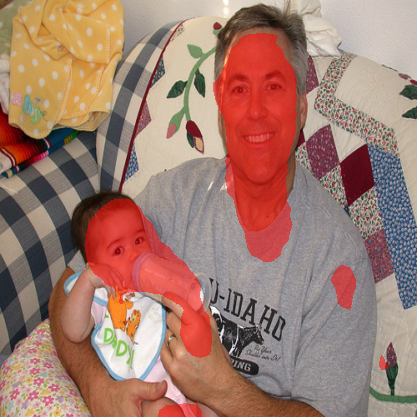} &
      \includegraphics[width=\linewidth]{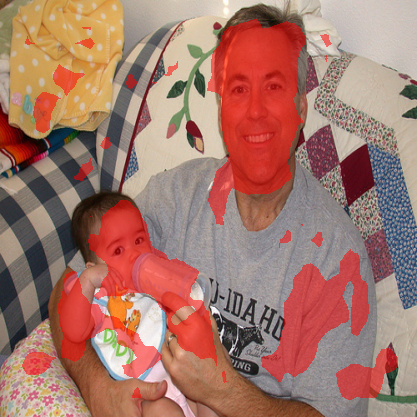} &
      \includegraphics[width=\linewidth]{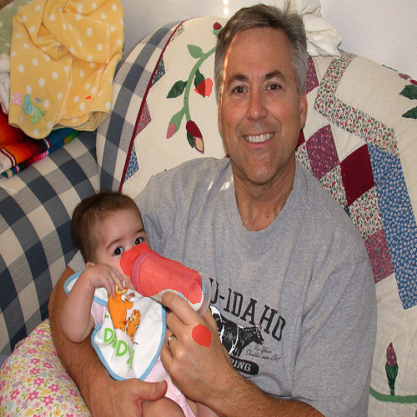} &
      \includegraphics[width=\linewidth]{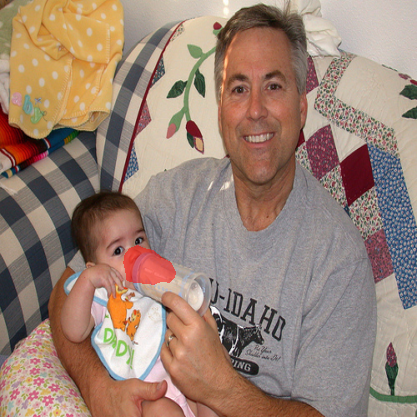} \\
      &
      0.00 \acrshort{iou} &
      0.09 \acrshort{iou} &
      0.06 \acrshort{iou} &
      0.55 \acrshort{iou} &
      0.32 \acrshort{iou} \\
    \end{tabularx}
    \captionsetup{justification=justified,margin=0cm}
    \caption{Qualitative Results: column (a) shows Query images with ground truth labels. The predictions of the baseline models are displayed in column (b) for MobileNet-V3-Large, column (c) VGG16, and column (d) for ResNet50. These include notable false-positive and false-negative predictions, revealing the challenges in accurately capturing certain object parts. In contrast, predictions from ensemble techniques configured with all three backbones demonstrate significant improvements. As shown in columns (e) Independent Voting and (f) Feature Volume Fusion, subject coverage is enhanced, compensating for the limitations observed for the individual baselines. Under each prediction we also report the \acrshort{iou} score achieved.}
    \label{fig:qualitative}
  \end{figure*}

Furthermore, Fig. \ref{fig:qualitative} presents a few qualitative examples, where the ground truth regions are shown alongside the predictions from individual baselines and ensembling strategies that combine all three backbones. A visual analysis of the segmentation masks exposes notable segmentation errors in the baseline predictions, which correspond to the lower performance figures in Tables \ref{tab_pascal} and \ref{tab_coco}. Crucially, both ensembling strategies significantly reduce the number of false positive predictions and improve the overall coverage of the subject. The enhanced quality of masks produced by both ensembling strategies is consistent with the superior numerical results presented in Tables \ref{tab_pascal} and \ref{tab_coco}.

\section{Implementation details}
The original implementation of PANet uses VGG16 as backbone, with weights pre-trained on ImageNet \cite{deng2009imagenet}. In addition to this default configuration, in our experiments, we explored the integration of alternative backbones in the PANet model: ResNet50, MobileNet-V3-Large, and their ensembled ablations. In all configurations, we initialised the model with weights learned from pre-training on ImageNet.

Throughout both the training and testing phases, we adhered to the methodology proposed by Wang et al. \cite{panet}. Input images were resized to $417 \times 417$ and augmentated via random horizontal flipping. End-to-end training was performed via stochastic gradient descent, with momentum set to 0.9 over 30,000 iterations. The learning rate was set 1e-3 and incrementally decreased by 0.1 every 10,000 iterations, while also applying a weight decay of 0.0005.

We relied  on the PyTorch framework for implementing our experiments\footnote{Code is redacted for anonymity and it will be released upon acceptance.}, 
building upon the PANet codebase shared by Wang et al. \cite{panet}. All experiments were run on an NVIDIA TITAN X and GTX 1080 Ti GPU with 12GB of memory.

\section{Conclusions}

In conclusion, experimental results have consistently highlighted the superior performance of Independent Voting and Feature Volume Fusion ensembling techniques over individual baselines. These results hint toward latent complementarities between embeddings extracted from different backbones. Crucially, these synergistic effects were found in a scenario where the pre-training set was kept fixed across trials and in the absence of learnable parameters in the embedding processing and mask prediction stage. As such they are only inherent to the choice of multiple backbones. 

Overall, this evidence builds a compelling case for appling ensembling to support \acrshort{fss} tasks, adopting a holistic approach that leverages different backbones. Findings from this paper can be exploited to simplify the process of backbone selection, as combining multiple backbones was found to be the preferable choice in all tested scenarios. 

The modular design proposed in this paper opens up opportunities to extend the study of ensembling strategies on different architectures and tasks. Future opportunities to extend this work lie in exploring embeddings derived from transformers and feature patches acquired through self-attention mechanisms \cite{vit,mae}. 

Another promising research avenue is the investigation of state-of-the-art models that operate without attention in a more resource-efficient setup, a particularly desirable feature in \acrshort{fss} settings - see, e.g., \cite{vim}. These future directions could further accelerate the progress on Domain Adaptation tasks that require robust \acrshort{fss} capabilities.

\bibliographystyle{ieeetr}
\bibliography{refs.bib}

\end{document}